\documentclass[10pt,twocolumn,letterpaper]{article}

\usepackage[pagenumbers]{cvpr} 

\usepackage{amsfonts}
\usepackage{amsmath}
\usepackage{multirow}
\usepackage{subcaption}
\usepackage{xcolor}
\definecolor{mygray}{gray}{0.6}
\newcommand{\graycell}[1]{\textcolor{mygray}{#1}}
\definecolor{mygray2}{RGB}{220,225,235}
\usepackage{colortbl}
\definecolor{mygreen}{RGB}{210,230,220}
\usepackage{soul}
\usepackage{booktabs}
\usepackage{pifont}

\definecolor{cvprblue}{rgb}{0.21,0.49,0.74}
\usepackage[pagebackref,breaklinks,colorlinks,allcolors=cvprblue]{hyperref}

\def\paperID{} 
\def\confName{CVPR}
\def\confYear{2026}

\title{PET-DINO: Unifying Visual Cues into Grounding DINO \\
with Prompt-Enriched Training}

\author{
Weifu Fu$^{1\dag}$\thanks{Equal contribution, \ $^\dag$Corresponding author.} \hfill
Jinyang Li$^{2*}$ \hfill
Bin-Bin Gao$^{1}$ \hfill
Jialin Li$^{3}$ \\
Yuhuan Lin$^{1}$ \hfill
Hanqiu Deng$^{1}$ \hfill
Wenbing Tao$^{2}$ \hfill
Yong Liu$^{1}$ \hfill
Chengjie Wang$^{1}$ \\
{\small $^1$YouTu Lab, Tencent \hfill
$^2$Huazhong University of Science and Technology \hfill
$^3$Kling Team, Kuaishou Technology} \\
\tt \small \{ryanwfu, danylgao, choasliu\}@tencent.com, \{jinyangli, wenbingtao\}@hust.edu.cn
}

\begin{document}

\maketitle
\begin{abstract}
Open-Set Object Detection (OSOD) enables recognition of novel categories beyond fixed classes but faces challenges in aligning text representations with complex visual concepts and the scarcity of image-text pairs for rare categories. This results in suboptimal performance in specialized domains or with complex objects. Recent visual-prompted methods partially address these issues but often involve complex multi-modal designs and multi-stage optimizations, prolonging the development cycle. Additionally, effective training strategies for data-driven OSOD models remain largely unexplored.
To address these challenges, we propose \textbf{PET-DINO}, a universal detector supporting both text and visual prompts. Our \textbf{A}lignment-\textbf{F}riendly \textbf{V}isual \textbf{P}rompt \textbf{G}eneration (\textbf{AFVPG}) module builds upon an advanced text-prompted detector, addressing the limitations of text representation guidance and reducing the development cycle. We introduce two prompt-enriched training strategies: \textbf{I}ntra-\textbf{B}atch \textbf{P}arallel Prompting (\textbf{IBP}) at the iteration level and \textbf{D}ynamic \textbf{M}emory-\textbf{D}riven Prompting (\textbf{DMD}) at the overall training level. These strategies enable simultaneous modeling of multiple prompt routes, facilitating parallel alignment with diverse real-world usage scenarios.
Comprehensive experiments demonstrate that PET-DINO exhibits competitive zero-shot object detection capabilities across various prompt-based detection protocols. These strengths can be attributed to inheritance-based philosophy and prompt-enriched training strategies, which play a critical role in building an effective generic object detector.
Project page: \href{https://fuweifuvtoo.github.io/pet-dino/}{https://fuweifuvtoo.github.io/pet-dino}.
\end{abstract}

\section{Introduction}
\label{sec:intro}

As a fundamental pillar of visual perception, object detectors play a crucial role in a wide range of real-world applications. However, conventional detection models~\cite{zhu2020deformable, fu2021lvis, zhang2022dino, li2024lors} are restricted to a fixed set of predefined categories, limiting their ability to handle the vast diversity of visual concepts. Open-Set Object Detection (OSOD) enables models to recognize and localize novel visual concepts unseen during training, thereby overcoming the limitations of traditional detectors. This advancement paves the way for more general and intelligent visual perception systems that can adapt to the evolving complexity of the real world.

Current mainstream open-vocabulary detection methods~\cite{gu2021open, li2022grounded, liu2024grounding, yao2024detclipv3} use large-scale data or vision-language models (VLMs) to align visual representations with pre-trained text encoders, enabling zero-shot generalization by capturing high-level category semantics.
Nevertheless, this paradigm still encounters significant challenges in truly open-world scenarios. First, text features often struggle to correspond effectively to visual concepts in specialized domains or for complex objects, making it difficult to accurately distinguish these categories. Second, the scarcity of image-text aligned samples for long-tail categories further impedes the model’s open-vocabulary performance.
In contrast, visual representations inherently contain rich information that extends beyond textual descriptions and are not constrained by cross-modal alignment. 
Consequently, recent research has explored the use of visual representations of objects as prompts (i.e., visual prompts) for detectors, leveraging their strong representational capacity to compensate for the limitations of text-only prompts.

Recent works, such as T-Rex2~\cite{jiang2024t}, aim to build models that accept both visual and textual prompts as input. 
By leveraging object visual representations to guide detection, T-Rex2 achieves strong zero-shot performance in both interactive and generic detection.
CP-DETR~\cite{chen2025cp} uses an efficient early cross-modal fusion encoder to enhance interactions between image features and category semantics, boosting generalization. Similarly, YOLOE~\cite{wang2025yoloe} achieves real-time detection, low cost, and strong generalization through efficient modality alignment and visual prompt encoding.

These approaches employ tightly coupled model architectures and rely on large-scale supervised training to realize open-set perception. However, such integrated solutions inevitably require complex multi-modal processing designs and multi-stage training processes, all of which prolong the development cycle. Furthermore, for models that are highly data-dependent, it is crucial to thoroughly explore and optimize training strategies, as this can significantly enhance data-driven effectiveness. Therefore, there remains considerable room for improvement in both model construction and training strategies within existing methods.

In terms of model construction, we aim to build upon an advanced text-prompted detector and concentrate on the development of efficient visual prompt schemes, thereby reducing the development cycle and harmonizing text and visual prompt routes.
Specifically, we propose an \textit{\textbf{Alignment-Friendly Visual Prompt Generation} (\textbf{AFVPG})} module that extracts visual prompt embeddings from enhanced image features and shares parameters with the text prompt branch for efficient visual cue injection. Within the detection transformer, the text-prompted process provides the model with fundamental high-level semantic understanding, while the visual prompts generated by our proposed module effectively align instance-level image representations.

Regarding training strategies, since visual prompts are derived from the input image itself, this limits the diversity of visual prompts and consequently affects the model’s open-set perception. This also makes it difficult to model cross-image visual prompts and class-level multi-image global visual prompts, thus restricting alignment with real-world usage scenarios. Moreover, offline pre-extraction of visual prompts is not applicable to iterative models during training. Therefore, we design training process-based strategies to address these issues.

To align with real-world scenarios, we design our training strategies to reflect three practical visual prompting modes: interactive exemplar-guided prompting, class-level global prompting, and cross-image prompting. Leveraging the iterative nature of training, we propose the first training strategies for the dual-modality prompted detector: \textit{\textbf{Intra-Batch Parallel Prompting} (\textbf{IBP})} and \textit{\textbf{Dynamic Memory-Driven Prompting} (\textbf{DMD})}. At the iteration level, IBP utilizes visual prompts from other images within the same batch as prompts for the current image, and aggregate prompts of the same category to also serve as visual prompts for the current image during training, thereby simultaneously modeling exemplar-guided, class-level global, and cross-image prompting. At the overall training level, DMD introduces a \textbf{\textit{Visual Cues Bank}} that dynamically stores extracted visual prompt representations. In each iteration, aggregated prompts sampled from the Visual Cues Bank are used to further enhance the class-level global prompting mode. These also improve classification performance.

Based on the above methods, we propose \textbf{PET-DINO}, a universal object detector compatible with both text and visual prompts, as well as the first corresponding training strategies. Analytical experiments show that our visual prompt generation scheme based on text-prompt-based pretraining achieves a higher upper bound compared to training with visual prompts alone, and that both training strategies are remarkably effective. Our method achieves significant zero-shot detection performance on COCO~\cite{lin2014microsoft}, LVIS~\cite{gupta2019lvis}, and ODinW~\cite{li2022elevater}, and delivers excellent results across various prompt-based detection routes. Furthermore, in-domain evaluations also show strong performance.

Our main contributions are summarized as follows:
\begin{itemize}
\item 
We propose an Alignment-Friendly Visual Prompt Generation module, endowing advanced text-prompted detector with enhanced perception capabilities for linguistically complex scenarios, while also preserving the performance of text-prompted detection and reducing the development cycle.
\item 
We propose two prompt-enriched training strategies. Intra-Batch Parallel Prompting facilitates the simultaneous modeling of multiple prompt routes, while Dynamic Memory-Driven Prompting employs a Visual Cues Bank to propagate and enrich visual prompts. These strategies collectively allow the model to handle diverse real-world scenarios in a parallel manner.
\item 
Our approach achieves competitive performance against mainstream visual-prompted open-set Object detectors across multiple benchmarks. Furthermore, experimental analysis confirms the advantages of text-prompted pre-training over training from scratch.
\end{itemize}

\section{Related Work}

\begin{figure*}[t]
\centering
\includegraphics[width=0.9\textwidth]{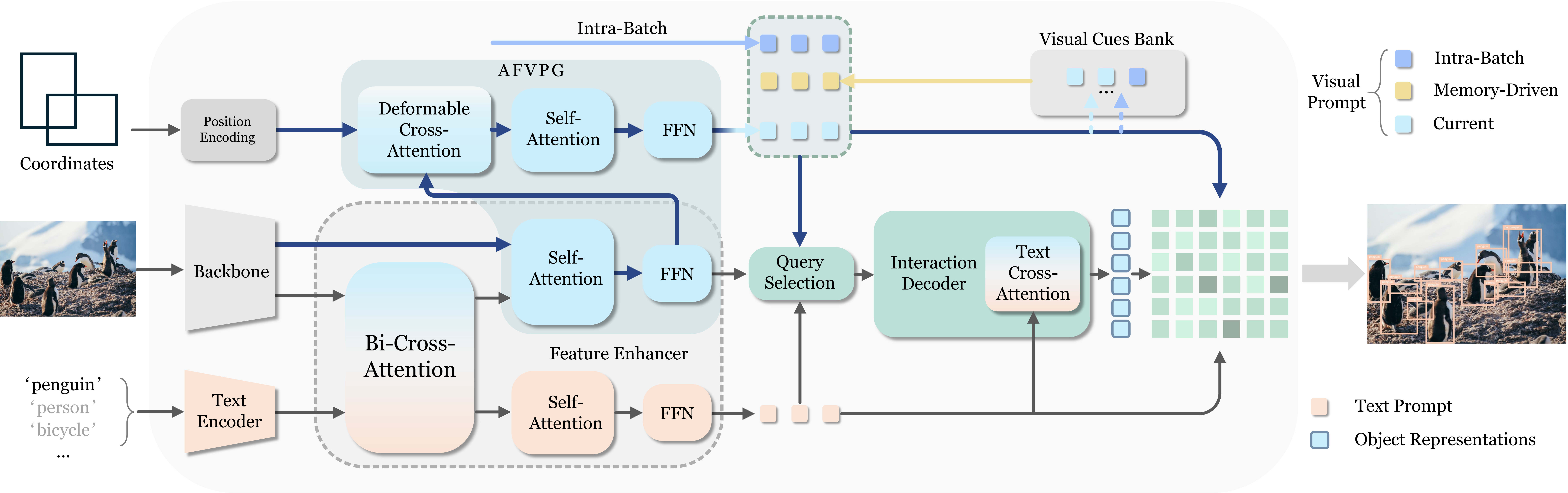} %
\caption{Overall architecture of PET-DINO. Input coordinates undergo a Visual Prompt Generation process, interacting with enhanced image features to obtain visual prompts. The text encoder creates text embeddings, which interact with image features in the Feature Enhancer module to produce text prompts. Both types of prompts guide the Query Selection Module, offering location priors for initial queries. These queries are then refined through decoder layers to predict objects and classifications.}
\label{method_fig1}
\vspace{-3mm}
\end{figure*}

\noindent\textbf{Text-prompted open-vocabulary object detection.}
Mainstream methods achieve zero-shot generalization by aligning visual features with text embeddings from pre-trained vision-language models, thereby capturing high-level category semantics.
ViLD~\cite{gu2021open} distills knowledge from the large-scale vision-language model CLIP~\cite{radford2021learning}. RegionCLIP~\cite{zhong2022regionclip} and DetCLIP~\cite{yao2022detclip} align language with image regions and leverage image-text pairs with pseudo boxes to enrich region-level knowledge for more generalized object detection. GLIP~\cite{li2022grounded} further unifies object detection and phrase grounding tasks, transforming detection into text-guided region localization. Grounding DINO~\cite{liu2024grounding} further proposes a feature enhancer and a cross-modality decoder to achieve denser fusion. GLEE~\cite{wu2024general}, using sentence-level text encoding, achieves object perception tasks without task-specific fine-tuning.
Moreover, text-prompted and long-tail perception paradigms have also been extended to segmentation and tracking~\cite{peng2025understanding,li2025ovtr,chen2024delving,chen2025cross,chen2025reamot}.

\noindent\textbf{Object detection with visual prompts.} 
While text prompts offer a generic description, certain objects can be challenging due to the potential mismatch between text descriptions and complex visual scenes. In such cases, visual prompts can serve as a complement to text prompts.
MQDet~\cite{xu2023multi} augments text queries with class-specific visual information from query images.
T-Rex~\cite{jiang2023t} is an interactive object counting model that can first detect and then count arbitrary objects.
T-Rex2~\cite{jiang2024t} proposes a general open-set object detection method that combines text and visual prompts and aligns the two prompt modalities via contrastive learning, effectively improving the model’s generality and zero-shot detection capability.
DINOv~\cite{li2024visual} enables segmentation and detection via visual in-context prompting, employing exemplars like masks and boxes to provide specific guidance.
CP-DETR~\cite{chen2025cp} proposes a unified detector that fuses text, visual, and optimized prompts, exploring more efficient cross-modal fusion mechanisms.
MI-Grounding~\cite{zhang2025just} introduces the “Image Prompt Paradigm”, which uses a few instance images as prompts to achieve open-set object detection and segmentation. 
YOLOE~\cite{wang2025yoloe} achieves real-time detection and segmentation of arbitrary objects across diverse open-world scenarios, supporting text, visual, or no prompts, balancing speed, accuracy, and flexibility.
\section{Method}

Built upon the text-prompted detector Grounding DINO~\cite{liu2024grounding}, PET-DINO introduces novel visual prompting and prompt-enriched training strategies, as shown in Figure~\ref{method_fig1}.

PET-DINO supports two routes: visual prompt route and textual prompt route. The input coordinates are processed by Alignment-Friendly Visual Prompt Generation (AFVPG) module to obtain visual prompts. For textual inputs, the text encoder produces text embeddings, which are further enhanced through interaction with image features in the Feature Enhancer module. Both types of prompt serve as guidance in the query selection module, providing location priors for the 900 initialized queries, which are then iteratively refined by a 6-layer interaction decoder to capture instance-level information for precise object detection.

We propose the first large-scale training strategies for visual prompt training, including Intra-Batch Parallel Prompting (IBP) and Dynamic Memory-Driven Prompting (DMD). 
This scheme aims to achieve alignment with visual prompt detection usage scenarios in a parallel manner, expand the discriminative range of categories, and enhance category sensitivity.
For optimization, the training objectives for our model are as follows:
\begin{equation}
    \mathcal{L} = \mathcal{L}_{L1} + \mathcal{L}_{GIoU} + \mathcal{L}_{alignment}
    \label{eq0}
\end{equation}
where $\mathcal{L}_{L1}$ and $\mathcal{L}_{GIoU}$ denote the localization losses, corresponding to the L1 loss and GIoU loss~\cite{rezatofighi2019generalized}, respectively. The term $\mathcal{L}_{alignment}$ refers to the focal loss~\cite{li2020generalized} used for classification.

\subsection{Prompt-guided Detection Routes}

\noindent\textbf{Inherited Text Prompt Route.} 
For textual inputs, category names are concatenated as input text to reformulate object detection as a grounding task.
The enhanced text features output from the feature enhancer module can be regarded as text prompt embeddings, which then serve as guidance in the query selection module.
The selected queries are fed into the cross-modality decoder, which incorporates a text cross-attention layer to inject text information into queries. Finally, text prompt embeddings are used for contrastive learning with instance-level representations output by the decoder, enabling open-vocabulary perception.

\noindent\textbf{Visual Prompt Routes.} 
We aim to develop two types of visual prompt routes with broad practical applications. The first, termed \textit{Exemplar-Guided Route}, enables the detector to interactively capture objects belonging to the same category as a given instance-level visual prompt. 
The second, referred to as \textit{Global-Concept Route}, aggregates various visual prompts from multiple images to generate a generic and comprehensive prompt for each category. By inputting global prompts for various categories, the model can achieve broad recognition across a wide range of object categories, thereby enabling comprehensive perception.

\subsection{Streamlined Visual Cue Injection}
To efficiently inject visual cues into the model while maintaining the performance of text prompt route, we design an Alignment-Friendly Visual Prompt Generation (AFVPG) module, which extracts visual prompt embeddings from the enhanced image features and incorporates a parameter-sharing mechanism with the text prompt branch.

Specifically, directly extracting instance-level visual prompt embeddings from the backbone's under-enhanced features is challenging. These features are denoted as $\boldsymbol{x_i} \in \mathbb{R}^{C_i \times H_i \times W_i}$, where $i \in \{1,2,...,L\}$ indexes the $L$ feature map layers.
Moreover, early interaction between visual prompts and image features may gradually widen the gap between them. To address these issues, we instead use enhanced image features $\boldsymbol{x'_i} \in \mathbb{R}^{C_i \times H_i \times W_i}$, which have been enhanced by deformable self-attention~\cite{zhu2020deformable} and feed-forward networks (FFN) in the Feature Enhancer.
The enhanced image features are obtained as follows:
\begin{equation}
\left\{\boldsymbol{x'_i}\right\}_{i=1}^L = \operatorname{FFN}(\operatorname{DeformSelfAttn}(\left\{\boldsymbol{x_i}\right\}_{i=1}^L))
\end{equation}

To ensure category balance and computational efficiency, we follow the inductive approach for each category's visual prompt as in T-Rex2~\cite{jiang2024t}.
Specifically, we initialize a learnable content embedding, which is then broadcast $K$ times for $K$ prompt boxes, denoted as $C \in \mathbb{R}^{K \times D}$. Additionally, we initialize an extra universal learnable content carrier $C' \in \mathbb{R}^{1 \times D}$ to aggregate features.
By concatenating $C$ and $C'$ and applying a linear layer for projection, we obtain the input query embedding $Q \in \mathbb{R}^{(K+1) \times D}$.

For each category, we first normalize the coordinates of $K$ prompt boxes. These normalized coordinates are then concatenated with a global normalized coordinate $[0.5, 0.5, 1, 1]$. The resulting $(K+1)$ coordinates are subsequently encoded using sine-cosine positional encoding and passed through a linear transformation, producing $Q_{pos} \in \mathbb{R}^{(K+1) \times D}$. The $Q$ and $Q_{pos}$ are generated as follows:
\begin{gather}
Q = \operatorname{Linear}\left(\operatorname{CAT}\left(\left[C; C'\right]\right)\right) \\
Q_{\text{pos}} = \operatorname{Linear}\left(\operatorname{PE}\left(b_1, \ldots, b_K, [0.5, 0.5, 1, 1]\right)\right)
\end{gather}

\begin{figure}[t]
\centering
\includegraphics[width=0.8\columnwidth]{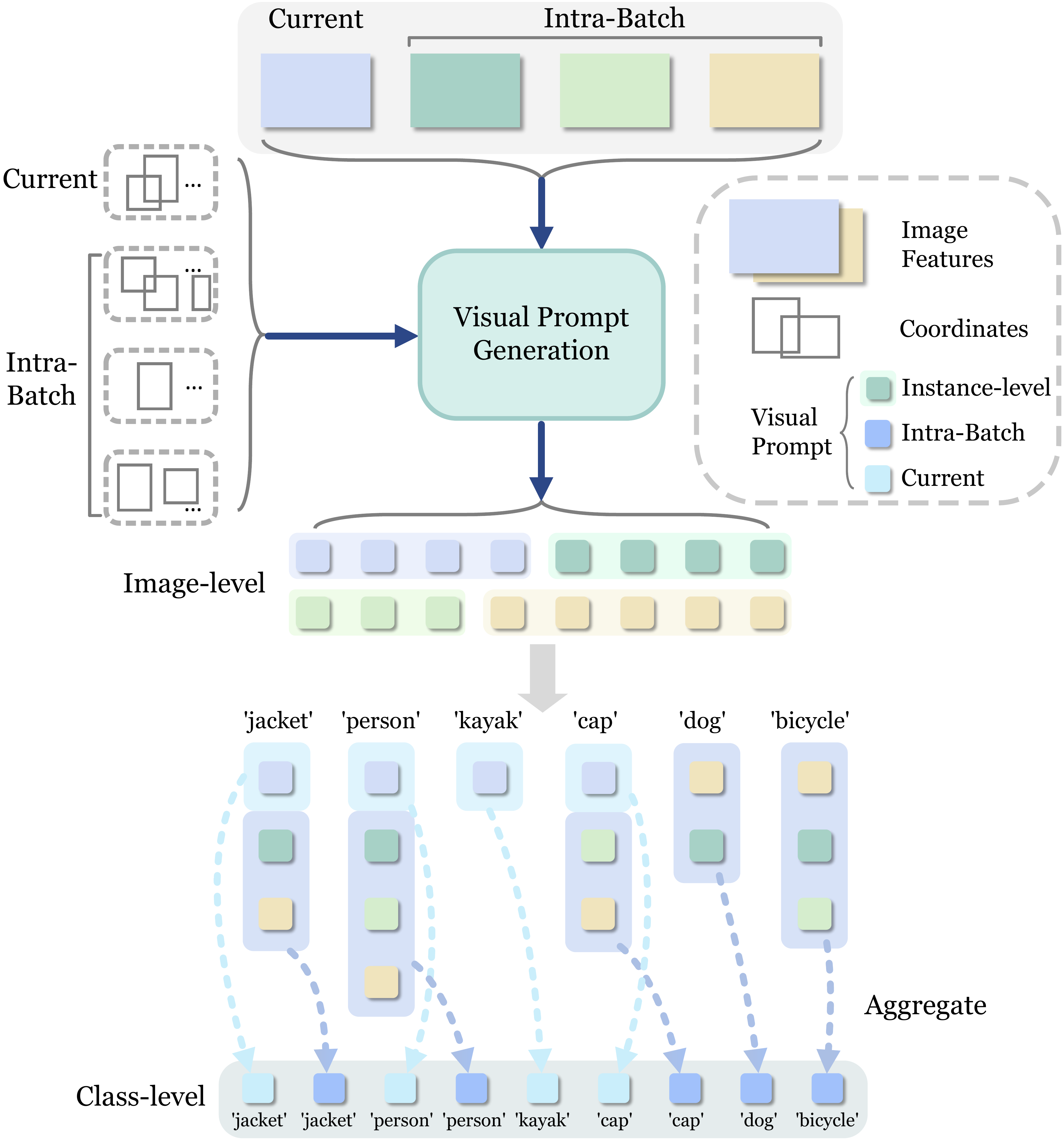} %
\caption{Intra-Batch Parallel Prompting Diagram. Image and their target coordinates within a batch are processed to generate image-level visual prompts. We incorporate visual prompts from other samples as additional prompts for the current image, aggregating those from the same category to form class-level visual prompts.}
\label{method_fig2}
\vspace{-3mm}
\end{figure}

The queries interact with the enhanced image features $\boldsymbol{x'_i}$ through multi-scale deformable cross-attention, enabling them to focus on instance-specific information of interest within the visual cues. Afterward, self-attention and a feed-forward network (FFN) are applied, allowing the universal content carrier to integrate information from other queries. This aggregation yields the global visual prompt embedding $V \in \mathbb{R}^{1 \times D}$. The process is formulated as follows.
\begin{equation}
{Q'} = \operatorname{MSDeformAttn}(Q, Q_{pos}, \left\{\boldsymbol{x'}_i\right\}_{i=1}^L)
\end{equation}
\begin{equation}
V=\operatorname{FFN}(\operatorname{SelfAttn}(Q'))[-1]
\end{equation}

We obtain the visual prompt embeddings of multiple categories and combine them as $[V_1, V_2, ..., V_c]$. These embeddings are obtained from enhanced image features $\boldsymbol{x'_i}$, thereby promoting better alignment with the instance information embedded within them. Ultimately, a grounding-like process guided by visual prompts can be achieved.
Notably, given the versatility of the deformable self-attention and FFN in the Feature Enhancer, we share them between visual prompt generation and text prompt detection.

\subsection{Intra-Batch Parallel Prompting}
We leverage the efficient parallelism of mini-batch training by treating the visual prompts corresponding to different samples within the same batch as negative samples for each other and aggregating them. This approach aims to break the compressed category discrimination space caused by single-image visual prompts and to achieve alignment with general usage scenarios, including both \textit{Exemplar-Guided Route} and \textit{Global-Concept Route}, in a parallel manner.

The Intra-Batch Parallel (IBP) Prompting is shown in Figure~\ref{method_fig2}.
A batch of images $[I_1, I_2, \ldots, I_N]$, where $N$ denotes the batch size, together with their corresponding coordinates, is fed into the visual prompt generation process. This process produces all image-level visual embeddings for each category, denoted as $V_c^j$, where $c$ represents the category id and $j$ represents the sample index in the batch. For category $c$, all visual embeddings of category $c$ within the batch are grouped together. This group is denoted as $\mathcal{V}_c=[ \ldots, V_c^j, V_c^{j+1}, \ldots]$, which contains $n$ elements, where $n \leq N$.

For the visual embeddings of category $c$ from non-current images $[I_1, \ldots, I_{i-1}, I_{i+1}, \ldots, I_N]$ in the batch, they are aggregated as follows:
\begin{equation}
V_c^{\text{batch}} = 
\begin{cases}
\frac{1}{n-1} \displaystyle\sum_{\substack{j \in \mathcal{J} \\ j \neq i}} V_c^j, & \text{if } i \in \mathcal{J} \\
\frac{1}{n} \displaystyle\sum_{j \in \mathcal{J}} V_c^j, & \text{if } i \notin \mathcal{J}
\end{cases}
\end{equation}
where $J$ is the set of indices of images in the batch that contain category $c$, $V_c^{batch}$ denotes the aggregated visual prompt from non-self samples. In this way, for category $c$ in image $I_i$, we obtain two visual prompts, denoted as $\{V_c^i, V_c^{batch}\}$.

Here, $V_c^i$ is derived from the self-image and is responsible for interactive detection during training (\textit{Exemplar-Guided Route}). In contrast, $V_c^{batch}$ is obtained by aggregating the embeddings from other images in the batch, which not only facilitates the optimization of cross-image visual prompt detection, but also enables class-level, multi-image global visual prompt detection (\textit{Global-Concept Route}). Thus, this design enables the alignment between training and general usage scenarios in a parallel fashion. Furthermore, when the current image does not contain category $c$, i.e., $V_c^i \notin \mathcal{V}_c$, Intra-Batch Parallel Prompting also enlarges the category discrimination space and enhances the category sensitivity of prompt-based detection.
In addition, when multiple datasets are involved in training, each batch is constructed by sampling images from only a single dataset to prevent semantic ambiguity arising from conflicting category definitions.

\subsection{Dynamic Memory-Driven Prompting}

While the IBP strategy offers a strong foundation, its capacity for generating diverse prompts across training iterations is limited. To address this, we introduce Dynamic Memory-Driven (DMD) Prompting, which is designed to further expand prompt diversity and drive additional performance gains in the \textit{Global-Concept Route} scenario.

We introduce a Visual Cues Bank as a carrier for visual prompt propagation during iterative training. We employ a dynamic update strategy, which offers superior adaptability to continually iterative networks compared to offline visual prompts. The Visual Cues Bank is updated with newly extracted visual prompt embeddings, including both current and intra-batch prompts, thus maintaining a continually iterative representation space for dynamic prompting. Specifically, for each category, we maintain a first-in-first-out (FIFO) queue of length $M$ to ensure prompt updates.
During each training iteration, we randomly select $d$ categories (including both negative and positive ones) from the dataset and sample their corresponding prompt embeddings from the Visual Cues Bank. 

For each category $c$, We extract the historical visual prompts $[\widetilde{V}_c^1, \widetilde{V}_c^2, \ldots, \widetilde{V}_c^M]$ from the Visual Cues Bank. To align with the \textit{Global-Concept Route} scenario, we further aggregate these embeddings to obtain a generic visual representation for category $c$:
\begin{equation}
V_c^{\mathrm{mem}} = \frac{1}{M} \sum_{k=1}^{M} \widetilde{V}_c^k
\end{equation}

Based on the above methods, our approach \textit{models three types of visual prompting in parallel}. For each category $c$ in image $i$, we obtain three distinct visual prompt embeddings, denoted as $\{V_c^i, V_c^{\mathrm{batch}}, V_c^{\mathrm{mem}}\}$. As an additional benefit, the Visual Cues Bank facilitates efficient training by providing diverse visual prompt embeddings, which enables contrastive learning between categories that rarely co-occur within the same image.
Since each dataset possesses a distinct category dictionary, we maintain separate Visual Cues Banks for each training dataset (e.g., Object365, OpenImages) to preserve semantic consistency within each domain.

\section{Experiments}

\begin{table*}[t]
\centering
\resizebox{1.0\textwidth}{!}{ 
\begin{small}
\begin{tabular}{lccc|c|c|l|c|c|c}
\toprule
\multirow{2}{*}{Method} & Open & Prompt & \multirow{2}{*}{Backbone} & \multirow{2}{*}{VLM sup.} & Data & \multirow{2}{*}{Training Data} & \multirow{2}{*}{COCO} & \multirow{2}{*}{LVIS} &  \multirow{2}{*}{ODinW35} \\ 
       & Source & Type  &    &    & Size &    &    &      &  \\ \midrule
T-Rex2~\cite{jiang2024t} & N & Visual-I & Swin-T & CLIP &3.1M & O365, OpenImages, HierText, CrowdHuman, SA-1B & 56.6 & 59.3 & 37.7 \\
CP-DETR-T~\cite{chen2025cp} & N & Visual-I & Swin-T & CLIP & 3.3M & O365, OpenImages, GoldG, V3Det & 61.8 & \underline{64.1} & \underline{41.0} \\
\rowcolor{mygray2} \textbf{PET-DINO} & Y & Visual-I & Swin-T & -  & 0.6M & O365 & \underline{64.0} & 61.8 & 38.8 \\ 
\rowcolor{mygray2} \textbf{PET-DINO} & Y & Visual-I & Swin-T & -  & 2.5M & O365, OpenImages, V3Det & \textbf{64.3} & \textbf{64.5} & \textbf{48.3} \\ \midrule
T-Rex2~\cite{jiang2024t} & N & Visual-I & Swin-L & CLIP  &3.1M  & O365, OpenImages, HierText, CrowdHuman, SA-1B & 58.5 & 62.5 & 39.7 \\
CP-DETR-L\textbf{*}~\cite{chen2025cp} & N & Visual-I & Swin-L & CLIP  & 3.3M & O365, OpenImages, GoldG, V3Det & \graycell{68.4} & \graycell{71.6} & \textbf{50.6} \\
\rowcolor{mygray2} \textbf{PET-DINO} & Y & Visual-I & Swin-L & -  & 0.6M & O365 & \underline{66.4} & \underline{64.3} & 39.0 \\ 
\rowcolor{mygray2} \textbf{PET-DINO} & Y & Visual-I & Swin-L & -  & 2.5M & O365, OpenImages, V3Det & \textbf{66.5} & \textbf{65.8} & \underline{49.7} \\ \bottomrule
\end{tabular}
\end{small}
}
\vspace{-1mm}
\caption{Comparison for zero-shot interactive object detection through visual prompts. Fixed AP is reported on LVIS \texttt{minival} set.
\textbf{Bold} and \underline{underline} denote the best and second-best results in each column, respectively.
\textbf{*} means the training set of the evaluation dataset was used for pre-training with text prompts, so this model is not in a zero-shot setting.
}
\vspace{-1mm}
\label{table_visual-I}
\end{table*}

\begin{table*}[t]
\centering
\resizebox{1.0\textwidth}{!}{
\begin{small}
\begin{tabular}{lccc|c|c|l|c|c|c}
\toprule
\multirow{2}{*}{Method} & Open & Prompt & \multirow{2}{*}{Backbone} & \multirow{2}{*}{VLM sup.} & Data & \multirow{2}{*}{Training Data} & \multirow{2}{*}{COCO} & \multirow{2}{*}{LVIS} &  \multirow{2}{*}{ODinW35} \\ 
       & Source & Type  &    &    & Size &    &    &       &  \\ \midrule
T-Rex2~\cite{jiang2024t} & N &  Visual-G & Swin-T & CLIP & 3.1M & O365, OpenImages, HierText, CrowdHuman, SA-1B & \underline{38.8} & \textbf{37.4} & \underline{23.6} \\ 
YOLOE-v11-S~\cite{wang2025yoloe} & Y & Visual-G & YOLO-v11-S & MobileCLIP & 1.4M & O365, GoldG & - & 26.2 & - \\
\rowcolor{mygray2} \textbf{PET-DINO} & Y & Visual-G & Swin-T & - & 0.6M & O365 & \textbf{40.3} & 29.6 & 20.4 \\  
\rowcolor{mygray2} \textbf{PET-DINO} & Y & Visual-G & Swin-T & - & 2.5M & O365, OpenImages, V3Det & 38.4 & \underline{31.5} & \textbf{25.5} \\ 
\midrule
T-Rex2~\cite{jiang2024t} & N & Visual-G & Swin-L & CLIP & 3.1M & O365, OpenImages, HierText, CrowdHuman, SA-1B & \underline{46.5} & \textbf{47.6} & \underline{27.8} \\
YOLOE-v11-L~\cite{wang2025yoloe} & Y & Visual-G & YOLOv11-L & MobileCLIP & 1.4M & O365, GoldG & - & 33.7 & - \\
\rowcolor{mygray2} \textbf{PET-DINO} & Y & Visual-G & Swin-L & - & 0.6M & O365 & \textbf{46.7} & 34.9 & 24.2 \\ 
\rowcolor{mygray2} \textbf{PET-DINO} & Y & Visual-G & Swin-L & - & 2.5M & O365, OpenImages, V3Det & 44.3 & \underline{35.7} & \textbf{28.6} \\
\bottomrule
\end{tabular}
\end{small}
}
\vspace{-1mm}
\caption{Comparison for zero-shot generic object detection through visual prompts. Fixed AP is reported on LVIS \texttt{minival} set. \textbf{Bold} and \underline{underline} denote the best and second-best results in each column, respectively.
}
\vspace{-1mm}
\label{table_visual-G}
\end{table*}

\subsection{Dataset settings}
We train our models using datasets of varying scales to facilitate comprehensive experimentation. For the small-scale setting, we use only the Object365 dataset~\cite{shao2019objects365}. In the large-scale setting, we further incorporate data from OpenImages~\cite{kuznetsova2020open} and V3Det~\cite{wang2023v3det}.
All datasets employed in our experiments are publicly available. Notably, our approach requires a significantly smaller amount of data compared to other methods such as T-Rex2\cite{jiang2024t} and CP-DETR\cite{chen2025cp}.
For evaluation, we report zero-shot detection performance on COCO~\cite{lin2014microsoft}, LVIS minival~\cite{gupta2019lvis, kamath2021mdetr}, and ODinW~\cite{li2022elevater}.

\subsection{Implementation Details}
We develop two model variants based on Swin-T~\cite{liu2021swin} and Swin-L. We use BERT-base~\cite{devlin2019bert} as the text encoder, and adopt the checkpoints from MM-Grounding-DINO~\cite{zhao2024open}.
We use AdamW~\cite{loshchilov2017decoupled} as the optimizer with a weight decay of 1e-4. All experiments are run on 8 GPUs with a total batch size of 32. During training, only the network modules related to visual prompt routes are updated with a learning rate of 1e-4, while the image backbone and other modules are frozen. Models are trained for 12 epochs, with the learning rate reduced by a factor of 10 at epochs 8 and 11. We use a cyclical training strategy: 8 iterations of visual prompt tuning followed by 1 iteration of text prompt tuning.
For Dynamic Memory-Driven Prompting, we set $M$ to 16. The hyperparameter $d$ is set to 40 for category-rich datasets such as Object365, OpenImages, and V3Det.

\subsection{Protocols and Metrics}
Our method aims to align model training with real-world application scenarios: \textit{Exemplar-Guided Route} and \textit{Global-Concept Route}. During testing, these can be evaluated using the Visual-I and Visual-G protocols~\cite{jiang2024t}, respectively.
Therefore, we use three different evaluation protocols under different routes.

In the Visual-I protocol, models operate under the \textit{Exemplar-Guided Route}. For each test image, we randomly select one ground-truth (GT) box for each category to obtain visual prompt. This protocol effectively reflects the core capabilities required for interactive object detection and has a wide range of applications, such as automatic annotation.
In the Visual-G protocol, models operate under the \textit{Global-Concept Route}. For each category, we randomly sample 16 images containing that category from the training set and extract their visual prompt embeddings offline. Then we compute the average embedding for each category and use these embeddings as input visual prompts.
In the Text protocol, all category names from the benchmark are used as text inputs to generate text prompts.
For all protocols, Average Precision (AP) is used as the evaluation metric. 

\begin{table*}[t]
\centering
\resizebox{0.8\textwidth}{!}{
\begin{small}
\begin{tabular}{lcc|c|c|c|c}
\toprule
\multirow{2}{*}{Method} & Prompt & \multirow{2}{*}{Backbone} & \multirow{2}{*}{Training Data} & \multirow{2}{*}{COCO} & \multirow{2}{*}{LVIS} &  \multirow{2}{*}{ODinW35} \\ 
       & Type  &    &     &    &       &  \\ \midrule
Grounding DINO~\cite{liu2024grounding} & Text & Swin-T & O365, GoldG, Cap4M & 48.4 & 27.4 & 22.3 \\
MM-Grounding-DINO~\cite{zhao2024open} & Text & Swin-T & O365, GoldG, V3Det & 50.6 & 40.5 & 21.4 \\ 
\rowcolor{mygray2} \textbf{PET-DINO} & Text & Swin-T & O365\textsuperscript{\textdagger} & 49.8 \textcolor{blue}{$^{ \downarrow 0.8}$} & 37.8 \textcolor{blue}{$^{ \downarrow 2.7}$} & 20.6 \textcolor{blue}{$^{ \downarrow 0.8}$} \\ \midrule
Grounding DINO~\cite{liu2024grounding} & Text & Swin-L & O365, OpenImages, GoldG & 52.5 & 33.9 & 26.1 \\
MM-Grounding-DINO~\cite{zhao2024open} & Text & Swin-L & O365V2, OpenImage, GoldG & 53.0 & 36.7 & 23.9 \\
\rowcolor{mygray2} \textbf{PET-DINO} & Text & Swin-L & O365\textsuperscript{\textdagger} & 54.0 \textcolor{red}{$^{ \uparrow 1.0}$} & 39.3 \textcolor{red}{$^{ \uparrow 2.6}$} & 23.1 \textcolor{blue}{$^{ \downarrow 0.8}$} \\ \bottomrule
\end{tabular}
\end{small}
}
\vspace{-1mm}
\caption{Comparison for zero-shot generic object detection through text prompts. Fixed AP is reported on LVIS \texttt{minival} set. \textsuperscript{\textdagger} means that a small number of iterations on Object365 are not for dedicated text-prompt training, but specifically to prevent the degradation of the inherited text-prompt performance.}
\label{Text_eval}
\vspace{-1mm}
\end{table*}

\subsection{Comparison with Generic Detector}

\noindent\textbf{Exemplar-Guided Visual Prompt Evaluation.}
As shown in Table \ref{table_visual-I}, with the Swin-T backbone and using only the Objects365 dataset, PET-DINO achieves state-of-the-art zero-shot performance on COCO, surpassing CP-DETR by 2.2 AP. This demonstrates the high efficiency of our prompt-enriched training strategies, which enables strong performance even with limited data. 
Notably, CP-DETR uses 5.5 times more training data (3.3M vs. 0.6M), further proving the advantage of our method in reducing training cost. When trained with larger-scale datasets, PET-DINO significantly outperforms previous methods across all benchmarks. Compared to CP-DETR-T, the performance of Visual-I on ODinW35 is improved from 41.0 AP to 48.3 AP. When using the Swin-L backbone and training on multiple datasets, PET-DINO achieves 65.8 AP in zero-shot evaluation on LVIS minival. These experimental results indicate that our approach shows clear superiority in interactive object detection.

\noindent\textbf{Global-Concept Visual Prompt Evaluation.}
The Visual-G evaluation results are shown in Table \ref{table_visual-G}. With the Swin-T backbone and only Objects365 for training, PET-DINO achieves 40.3 AP on COCO, outperforming T-Rex2 by 1.5 AP. When trained with multiple datasets, Visual-G achieves 25.5 AP on ODinW35, surpassing T-Rex2 by 1.9 AP. However, on LVIS-minival, PET-DINO shows only limited performance compared to T-Rex2. This may be attributed, on one hand, to the limited long-tail category performance of our inherited pre-trained model~\cite{zhao2024open}, and on the other hand, to T-Rex2’s utilization of the open-vocabulary capabilities of the vision-language model CLIP~\cite{radford2021learning} and the SA-1B dataset produced by the data engine, which covers more diverse scenarios. When using the Swin-L backbone with Objects365, PET-DINO achieves state-of-the-art zero-shot performance on COCO with 46.7 AP. These results demonstrate the effectiveness of our model for generic object detection with visual prompts.

\begin{table}[t]
\centering
\resizebox{\columnwidth}{!}{%
\begin{small}
\begin{tabular}{lcc|c|c}
\toprule
\multirow{2}{*}{Method} & Prompt & \multirow{2}{*}{Backbone} & \multirow{2}{*}{Training Data} & \multirow{2}{*}{COCO} \\
       & Type  &   &        &   \\ \midrule
X-Decoder T~\cite{zou2023generalized} & Text & Focal-T &  COCO+CC3M+... & 43.6 \\
OpenSeed T~\cite{zhang2023simple} & Text & Swin-T &  COCO+O365 & 52.0 \\
OpenSeed L~\cite{zhang2023simple} & Text & Swin-L &  COCO+O365 & 58.2 \\ \midrule
MI Grounding-S~\cite{zhang2025just} & Image & ViT-L &  COCO+LVIS & 54.7 \\ \midrule
DINOv~\cite{li2024visual} & Visual-G & Swin-T &  COCO+SA-1B & 45.2 \\
DINOv~\cite{li2024visual} & Visual-G & Swin-L &  COCO+SA-1B & 54.2 \\
\rowcolor{mygray2} \textbf{PET-DINO} & Visual-G & Swin-T &  COCO+LVIS & \textbf{50.0} \\
\rowcolor{mygray2} \textbf{PET-DINO} & Visual-G & Swin-L &  COCO+LVIS & \textbf{56.3} \\
\bottomrule
\end{tabular}
\end{small}}
\vspace{-1mm} 
\caption{In-domain evaluation on COCO. We evaluate all models on COCO \texttt{val} set with the box AP reported.
}
\label{supervised} 
\end{table}

\begin{table}[t]
\centering
\resizebox{\columnwidth}{!}{%
\begin{tabular}{cc|c|c|c}
\toprule
T-Rex2 VP encoder & AFVPG & \begin{tabular}{@{}c@{}}COCO\\ Text\end{tabular} & \begin{tabular}{@{}c@{}}COCO\\ Visual-I\end{tabular} & \begin{tabular}{@{}c@{}}COCO\\ Visual-G\end{tabular} \\ \midrule
\checkmark &  & 49.6 & 59.2 & 37.6  \\ 
\rowcolor{mygreen} & \checkmark & 49.8 & 64.0 \textcolor{red}{$^{ \uparrow 4.8}$} & 40.3 \textcolor{red}{$^{ \uparrow 2.7}$} \\
\bottomrule
\end{tabular}
}
\caption{Ablation study of visual prompt generation. AFVPG indicates Alignment-Friendly Visual Prompt Generation and yields consistent improvements on both Visual-I and Visual-G protocols.}
\label{table_VPG}
\vspace{-1mm}
\end{table}

\begin{table}[t]
\centering
\resizebox{\columnwidth}{!}{%
\begin{tabular}{c|ccc|c|c|c}
\toprule
Row & AFVPG & IBP & DMD & \begin{tabular}{@{}c@{}}COCO\\ Text\end{tabular} & \begin{tabular}{@{}c@{}}COCO\\ Visual-I\end{tabular} & \begin{tabular}{@{}c@{}}COCO\\ Visual-G\end{tabular} \\ \midrule
0 & & & & 50.6 & - & -  \\ 
1 & \checkmark & & & 49.7 & 67.0 & 12.5  \\
2 & \checkmark & & \checkmark & 49.8 & 63.5 & 24.7  \\
3 & \checkmark & \checkmark & & 49.6 & 63.2 & 37.2  \\
\rowcolor{mygreen} 4 & \checkmark & \checkmark & \checkmark & 49.8 & 64.0 & 40.3  \\ \bottomrule
\end{tabular}
}
\caption{Ablation study of PET-DINO. IBP and DMD denote Intra-Batch Parallel prompting and Dynamic Memory-Driven prompting, respectively.}
\label{ablation_overall}
\vspace{-1mm}
\end{table}

\noindent\textbf{Text Prompt Evaluation.}
\label{sec:text-prompt-route-eval}
In this section, we evaluate the model’s performance under text prompts. Given that PET-DINO shares network components across text and visual prompt route, we examine the impact of this joint training paradigm on text prompt performance.
As shown in Table \ref{Text_eval}, PET-DINO with a Swin-T backbone shows slightly lower performance compared to its predecessor, MM-Grounding-DINO~\cite{zhao2024open}. This is expected due to the limited tuning of text prompt modules intended to preserve the original model's capabilities. Crucially, despite this constraint, PET-DINO's text prompt performance is not only well preserved but also enhanced with a more powerful backbone. Specifically, PET-DINO with a Swin-L backbone achieves improvements of 1.0 AP and 2.6 AP on COCO and LVIS \texttt{minival}, respectively.
These results indicate that the shared network has enough capacity to accommodate the joint training of both routes. Furthermore, we speculate that training with visual prompts may facilitate the enhancement of image features for the text prompt route.

\subsection{In-domain Evaluation on COCO} 
Table \ref{supervised} presents a comparison with several state-of-the-art universal models, all of which are trained on the COCO dataset. X-Decoder~\cite{zou2023generalized} and OpenSeed~\cite{zhang2023simple} are evaluated using the text prompt protocol, while DINOv~\cite{li2024visual} and PET-DINO adopt the Visual-G protocol, and MI Grounding-S~\cite{zhang2025just} utilizes the image prompt paradigm. With the Swin-L backbone, PET-DINO achieves a +2.1 AP improvement over DINOv. Remarkably, with only a few visual prompts per category, PET-DINO attains results comparable to well-established text-prompted open-set models. For instance, the performance gap between PET-DINO (Swin-L) and OpenSeed (Swin-L) is merely 1.9 AP. These results collectively underscore the strong effectiveness of PET-DINO and the robust performance of its visual prompt route in in-domain scenarios.

\subsection{Ablation Study}

To save computational resources, all ablation studies are conducted using a Swin-T backbone and are trained on the Object365 dataset.

\noindent\textbf{Effectiveness of AFVPG.} To demonstrate the effectiveness of the proposed Alignment-Friendly Visual Prompt Generation module, we replace it with T-Rex2 visual prompt encoder. As shown in Table \ref{table_VPG}, the visual prompt detection capability provided by the T-Rex2 visual prompt encoder proves to be suboptimal. In contrast, our AFVPG module achieves notable improvements (\textbf{+4.8} AP on Visual-I and \textbf{+2.7} AP on Visual-G), indicating that extracting instance-level visual prompt embeddings from the enhanced features is more effective than using the backbone's under-enhanced features, and since the AFVPG module shares parameters with the text prompt branch, the global high-level semantic representations from the text branch also contribute to the visual prompt learning, and further proving the effectiveness of the inheritance strategy.

\noindent\textbf{Overall Ablation Study of PET-DINO.}
In this section, we validate the effectiveness of our visual prompt generation scheme and the two prompt-enriched training strategies. 
As shown in Table \ref{ablation_overall}, Row 0 indicates the base model MM-Grounding-DINO, which is a text-prompted model that does not support visual prompt detection.
In Row 1, the incorporation of the Alignment-Friendly Visual Prompt Generation (AFVPG) module extends MM-Grounding-DINO to support visual prompts. Although the Visual-G metric remains low at this stage, the model achieves a notable Visual-I performance of \textbf{67.0} AP. 
Row 2 and Row 3 incorporate Dynamic Memory-Driven Prompting (DMD) and Intra-Batch Parallel Prompting (IBP), respectively, both of which improve Visual-G performance. In particular, Row 3 substantially increases the Visual-G metric from 12.5 AP to 37.2 AP (\textbf{+24.7} AP). Although Visual-I sees a slight decrease, we argue that this reflects a beneficial shift in the model’s focus: during inference, it relies less on replicating specific instances and more on recognizing the generalizable class patterns they represent.
Row 4 incorporates both AFVPG and two prompt-enriched training strategies, further boosting the Visual-G metric by \textbf{+3.1} AP to 40.3 AP and the Visual-I metric to 64.0 AP. Compared with Row 1, The adoption of these two prompt-enriched training strategies enables comprehensive parallel optimization, resulting in a highly efficient and unified training framework.

\subsection{Analysis of Text-Prompted Pre-training}
Our approach adopts an inheritance-based strategy, differing from multi-stage development methods. We build upon an advanced text-prompted detector, equipping it with the perceptual advantages of visual cues, while also reducing training costs and development cycles, and preserving the performance of text-prompted detection. In this section, we investigate the effectiveness and the impact of inheriting a pre-trained text-prompted detector.

\begin{figure}[t]
    \centering
    \begin{subfigure}[b]{0.47\columnwidth}
        \centering
        \includegraphics[width=\linewidth]{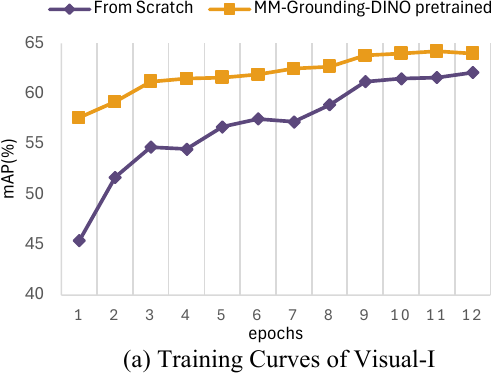}
    \end{subfigure}%
    \hspace{0.02\columnwidth}
    \begin{subfigure}[b]{0.49\columnwidth}
        \centering
        \includegraphics[width=\linewidth]{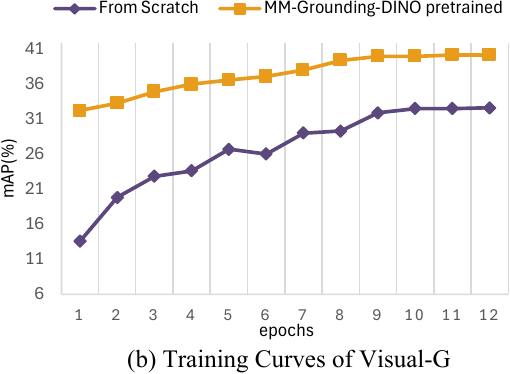}
    \end{subfigure}
    \caption{Comparison between training from scratch and inheriting the pre-trained model. The models are trained on O365 and evaluated on COCO \texttt{val} set.}
    \label{figure_transfer}
\end{figure}

We conducted comparative experiments between training a model from scratch and inheriting a pre-trained model. The results are presented in Table~\ref{table_transfer}. Our experiment demonstrates that initializing with a well-established text-prompted model as the pre-trained model yields superior performance compared to training from scratch. Specifically, the Visual-G protocol exhibits a significant performance improvement of \textbf{+7.6} AP. 
As illustrated in Figure \ref{figure_transfer}, after 12 epochs, the performance of the model trained from scratch has already stabilized, whereas inheriting the pre-trained MM-Grounding-DINO achieves a higher upper bound.
This indicates that text-prompted pre-training excels at leveraging global high-level semantic representations, which helps our model understand high-level category semantics, especially in the \textit{Global-Concept Route}.

\begin{table}[t]
\centering
\resizebox{\columnwidth}{!}{%
\begin{small}
\begin{tabular}{c|cc|c|c}
\toprule
\multirow{2}{*}{Method} & \multicolumn{2}{|c|}{Training Data}   & COCO  & COCO    \\  
       &MM-Grounding-DINO & PET-DINO  & visual-I    & visual-G \\ \midrule
PET-DINO-T     & \multirow{2}{*}{-}    &   \multirow{2}{*}{O365}   & \multirow{2}{*}{62.1}         &   \multirow{2}{*}{32.7}      \\
(from scratch) &&&& \\ \midrule
PET-DINO-T     & \multirow{2}{*}{O365, GoldG, V3Det} & \multirow{2}{*}{O365}  & \multirow{2}{*}{\textbf{64.0}}  & \multirow{2}{*}{\textbf{40.3}}         \\
(from pre-trained) &&&& \\
\bottomrule
\end{tabular}
\end{small}}
\caption{With or without inheriting the pre-trained model. All models are trained with Swin-T.}
\label{table_transfer}
\end{table}

\section{Conclusion}
In this paper, we propose PET-DINO, a versatile object detection framework that supports both text and visual prompting by extending the Grounding DINO architecture.
We begin by introducing the Alignment-Friendly Visual Prompt Generation (AFVPG) module, which enhances the model's sensitivity to complex and domain-specific visual cues and maintains text-prompting functionality. Additionally, we introduce two prompt-enriched training strategies, Intra-Batch Parallel Prompting (IBP) and Dynamic Memory-Driven Prompting (DMD), for visual prompt learning. These strategies align with diverse real-world scenarios and enhance overall performance by simulating a dense prompt environment during training.
PET-DINO achieves significant zero-shot detection performance across multiple benchmarks and experimental analysis demonstrates that text-prompt pre-training elevates the performance ceiling over training from scratch. This work adopts an inheritance-based philosophy and develops prompt-enriched training strategies to meet diverse practical scenarios. We hope this work can provide new insights for generic object detection.

{
    \small
    \bibliographystyle{ieeenat_fullname}
    \bibliography{main}
}

\clearpage
\appendix 
\setcounter{table}{0}
\setcounter{figure}{0}
\setcounter{equation}{0}

\clearpage
\setcounter{page}{1}
\maketitlesupplementary

\section{More Implementation Details}
\label{sec:More Implementation Details}

\subsection{Detailed Illustration of DMD}
\begin{figure}[ht]
\centering
\includegraphics[width=1.0\columnwidth]{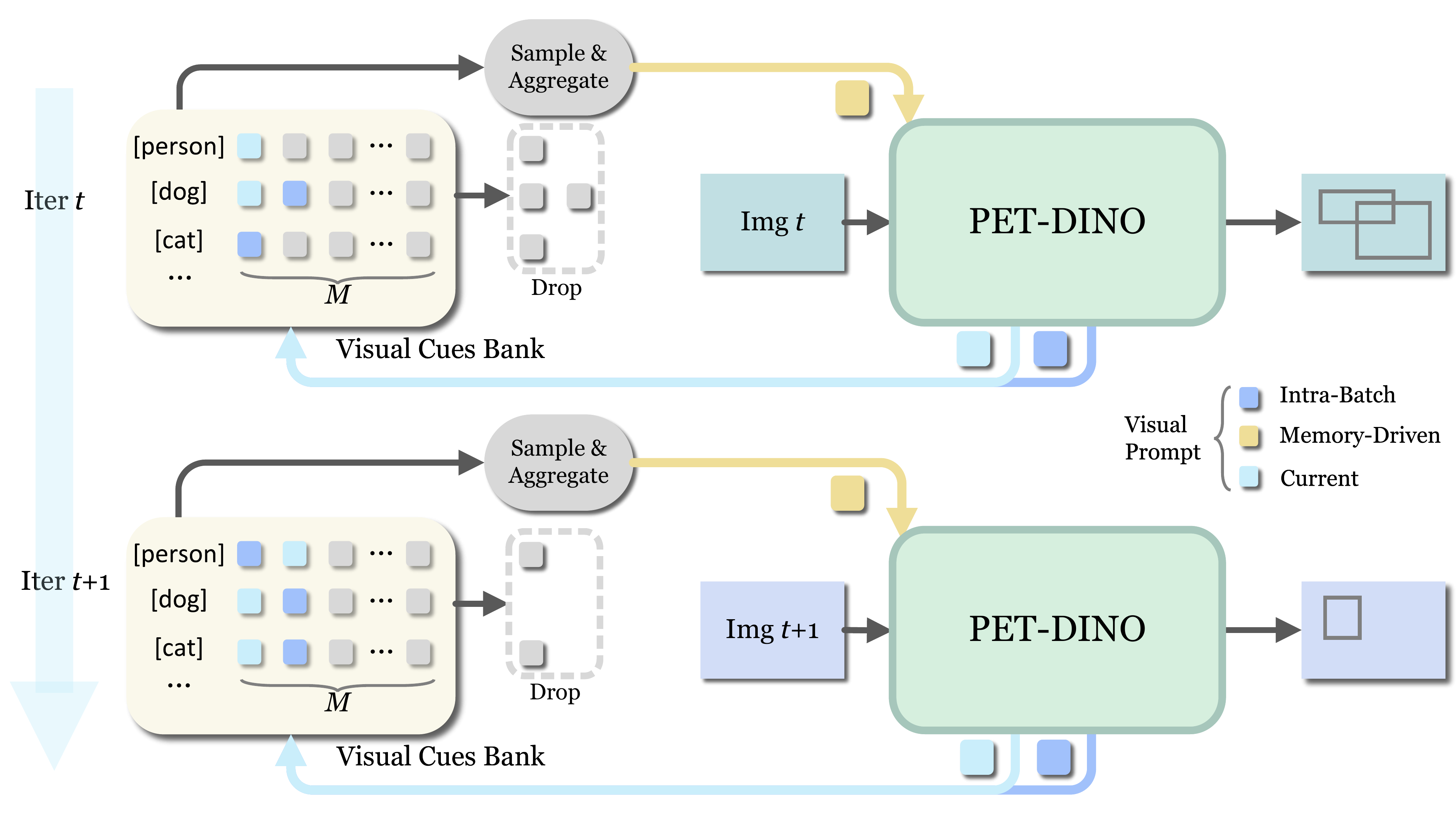} %
\caption{Dynamic Memory-Driven Prompting Diagram. During each iteration, the Visual Cues Bank updates its stored prompts with visual prompts used by PET-DINO, while PET-DINO utilizes the enriched prompts from the bank to improve training.}
\label{method_fig_a2}
\vspace{-3mm}
\end{figure}

Here, we further provide a detailed illustration of Dynamic Memory-Driven (DMD) Prompting, as shown in Figure~\ref{method_fig_a2}. At iteration \textbf{\textit{t}}, n categories are sampled from the Visual Cues Bank. For each sampled category, the corresponding stored visual prompts are aggregated to generate a memory-driven prompt. These memory-driven prompts, together with the current prompts and intra-batch prompts, are used in parallel to guide the model for detection, enabling parallel optimization, enhancing alignment with application scenarios, and strengthening open-category recognition. After participating in the guidance process, the current prompts and intra-batch prompts are stored by category to update the Visual Cues Bank, thereby enriching the visual cues for iteration \textbf{\textit{t}}+1 and significantly reducing computational overhead. To ensure that stored visual prompts remain compatible with the evolving network, the Visual Cues Bank keeps up to $M$ prompt embeddings per category during training, discarding the oldest when new ones are added.

\subsection{More Training Details}
Both the matching cost computation and the final loss calculation incorporate classification losses, box L1 losses, and GIoU losses. For classification, we employ a contrastive loss between predicted objects and prompt embeddings to align the prompt representations with the implicit information in the image. Specifically, we compute the dot product between each output query and the prompt embeddings to obtain logits for each category and then apply focal loss to each logit. For Hungarian matching, we assign weights of 2.0, 5.0, and 2.0 to the classification, L1, and GIoU costs, respectively. The corresponding loss weights are 1.0, 5.0, and 2.0 in the final loss calculation.

We use automatic mixed precision for training. For image augmentation, we adopt standard techniques used in DETR-like methods, including multi-scale training and random flipping. Following DINO, we employ contrastive denoising training (CDN) to stabilize training and accelerate convergence.

\section{Visualization Analysis of Proposed Modules}
\subsection{Feature correlation analysis of AFVPG}
\begin{figure}[t]
\centering
\includegraphics[width=1.0\columnwidth]{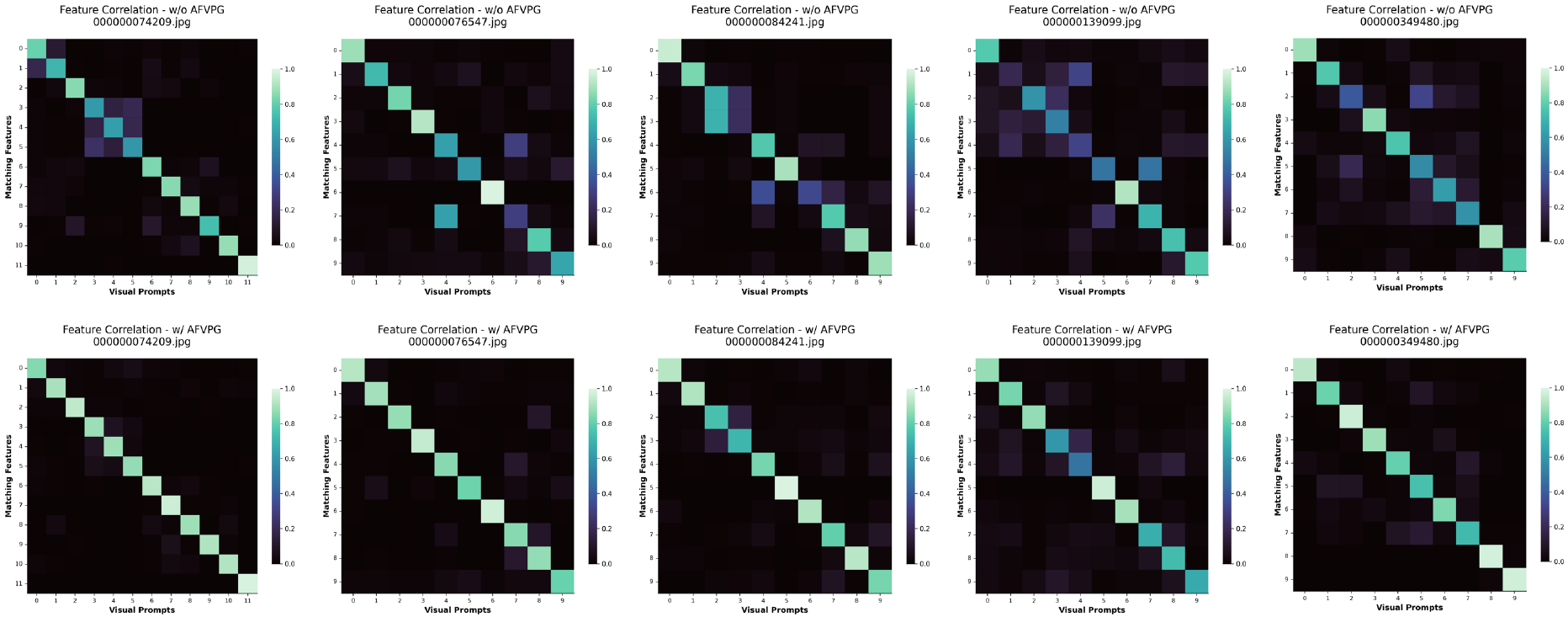}
\caption{Feature correlation analysis between visual prompts and instance-level image features showing the impact of AFVPG.}
\label{AFVPG_feture_analysis}
\end{figure}

To visually demonstrate the effectiveness of AFVPG, we provide a feature correlation analysis as shown in Figure~\ref{AFVPG_feture_analysis}. We compute the cosine similarity, followed by Softmax normalization, between all candidate prompts and the image feature most similar to each visual prompt.

As illustrated, AFVPG (bottom) markedly enhances diagonal responses and suppresses off-diagonal interference, effectively mitigating the cross-category ambiguity observed in the model without AFVPG (top). This intrinsic alignment facilitates precise query selection, yielding substantial performance gains of 4.8 AP on visual-I via this simple yet effective design.

\subsection{t-SNE visualization of IBP and DMD}

\begin{figure}[t]
\centering
\includegraphics[width=1.0\columnwidth]{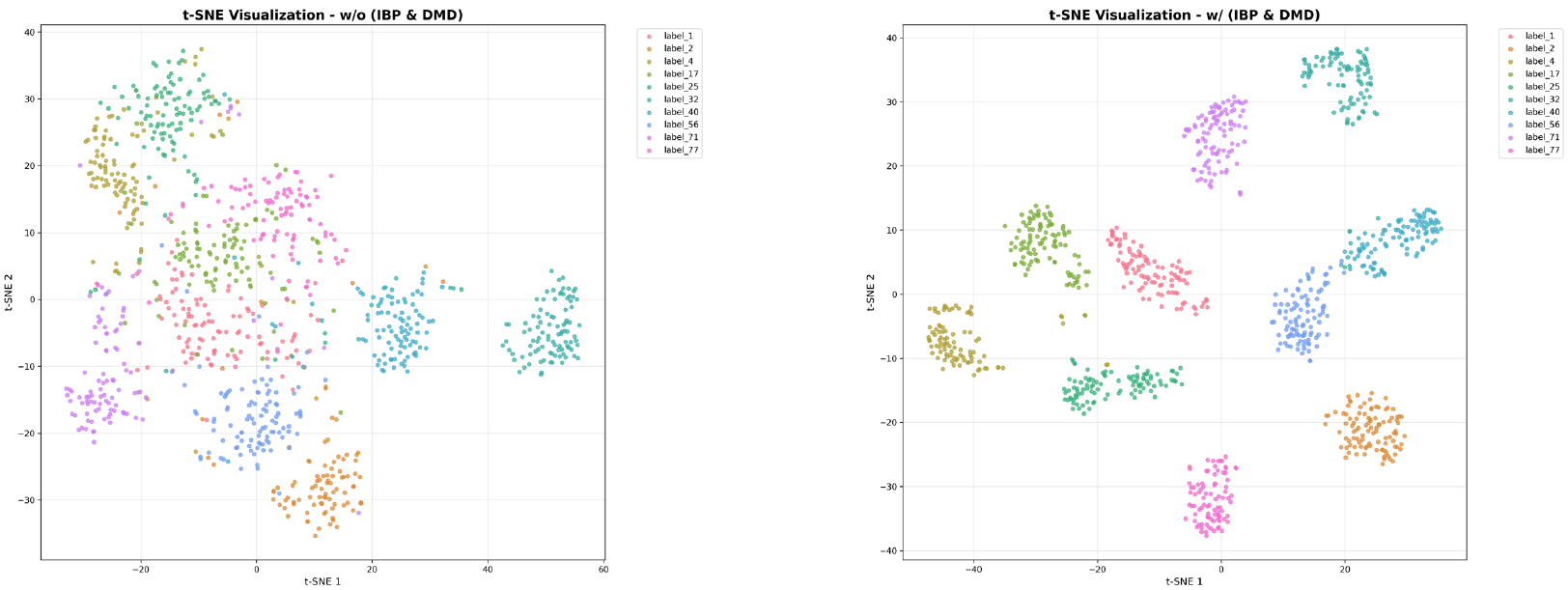}
\caption{t-SNE visualization of visual prompt features showing the impact of IBP and DMD.}
\label{IBP_DMD_tsne}
\end{figure}

To reveal the underlying reasons for the performance gains in Table 6, we conducted a t-SNE visualization on the visual prompt features of 10 randomly selected categories from the COCO dataset.

As shown in Figure~\ref{IBP_DMD_tsne}, without IBP and DMD (left), the visual prompt features are scattered without clear boundaries between categories. In contrast, incorporating IBP and DMD (right) leads to tighter intra-class aggregation and clearer inter-class separation.
This highly discriminative semantic space ensures robust category differentiation in complex multi-category scenarios, serving as the primary driver of performance boost.

\section{Visualizations}
\label{sec:Visualizations}
In this section, we comprehensively showcase the capabilities of PET-DINO across diverse scenarios by leveraging various types of prompts.

\begin{figure*}[ht]
\centering
\includegraphics[width=1.0\linewidth]{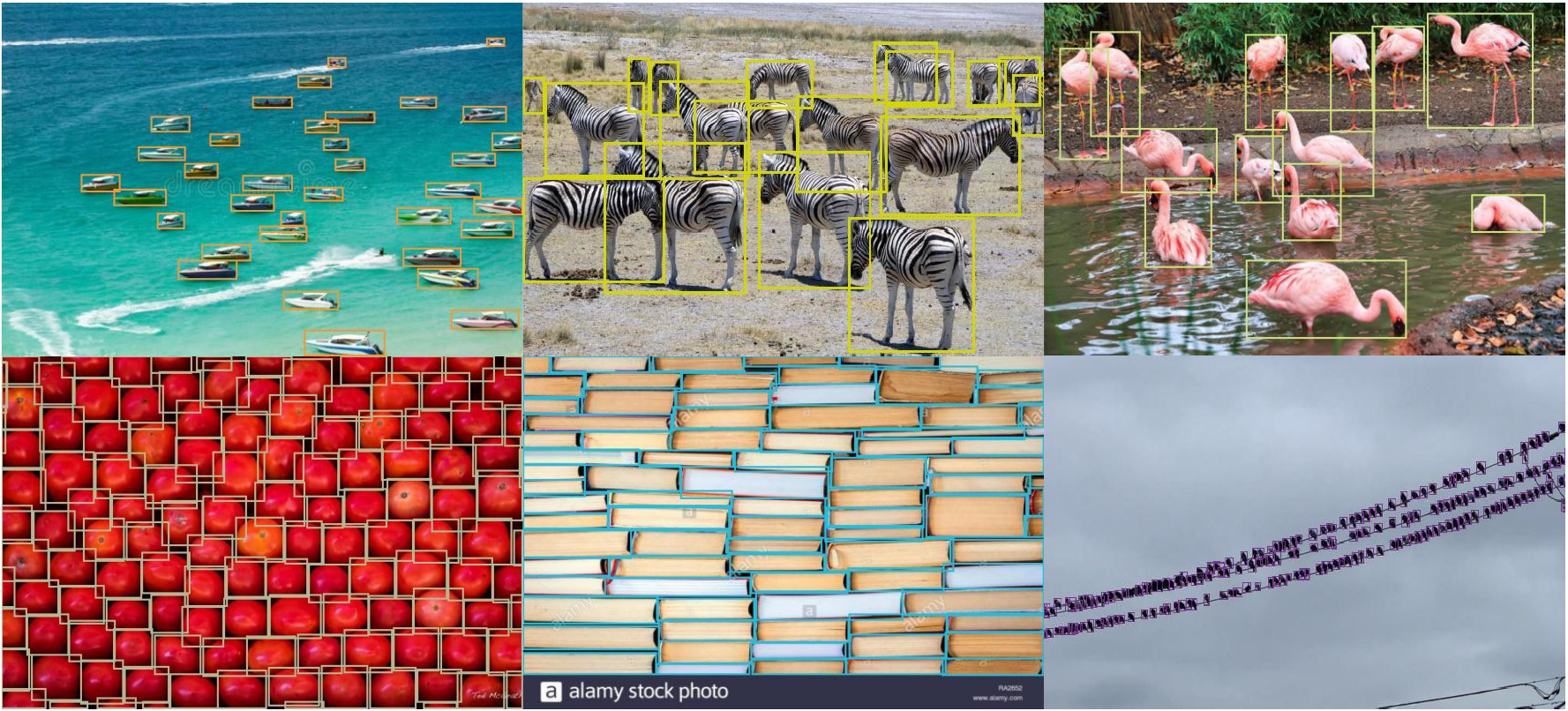} %
\caption{Zero-shot detection visualizations of \textbf{PET-DINO} on \textbf{interactive visual} prompt-based detection in \textbf{single}-category \textbf{dense} object scenarios.}
\label{visual_I_vis_single_dense_object}
\end{figure*}

\begin{figure*}[ht]
\centering
\includegraphics[width=1.0\linewidth]{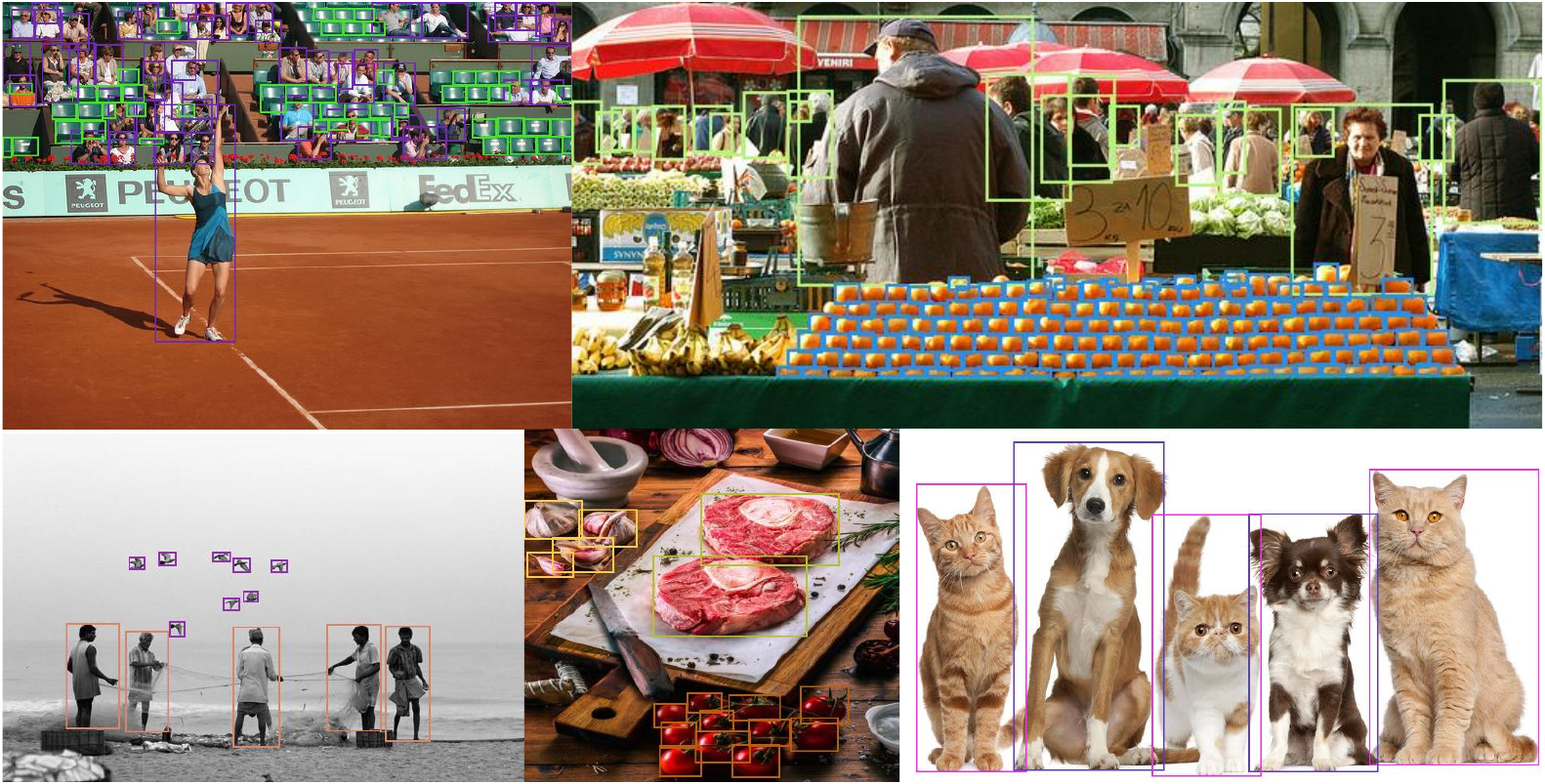} %
\caption{Zero-shot detection visualizations of \textbf{PET-DINO} on \textbf{interactive visual} prompt-based detection in \textbf{multi}-category \textbf{dense} object scenarios.}
\label{visual_I_vis_multi_objects}
\end{figure*}

We evaluated zero-shot detection performance in dense object scenarios using PET-DINO in the interactive visual prompt detection mode. As shown in Figure~\ref{visual_I_vis_single_dense_object}, PET-DINO demonstrates excellent performance in single-category scenes. Furthermore, as illustrated in Figure~\ref{visual_I_vis_multi_objects}, PET-DINO maintains strong performance in multi-category scenarios, accurately distinguishing between different categories. These results demonstrate that our prompt generation and training strategies enable the model to perform well in complex and dense detection scenarios, underscoring the strong potential of PET-DINO for real-world applications like automatic annotation and object counting.

\begin{figure*}[ht]
\centering
\includegraphics[width=1.0\linewidth]{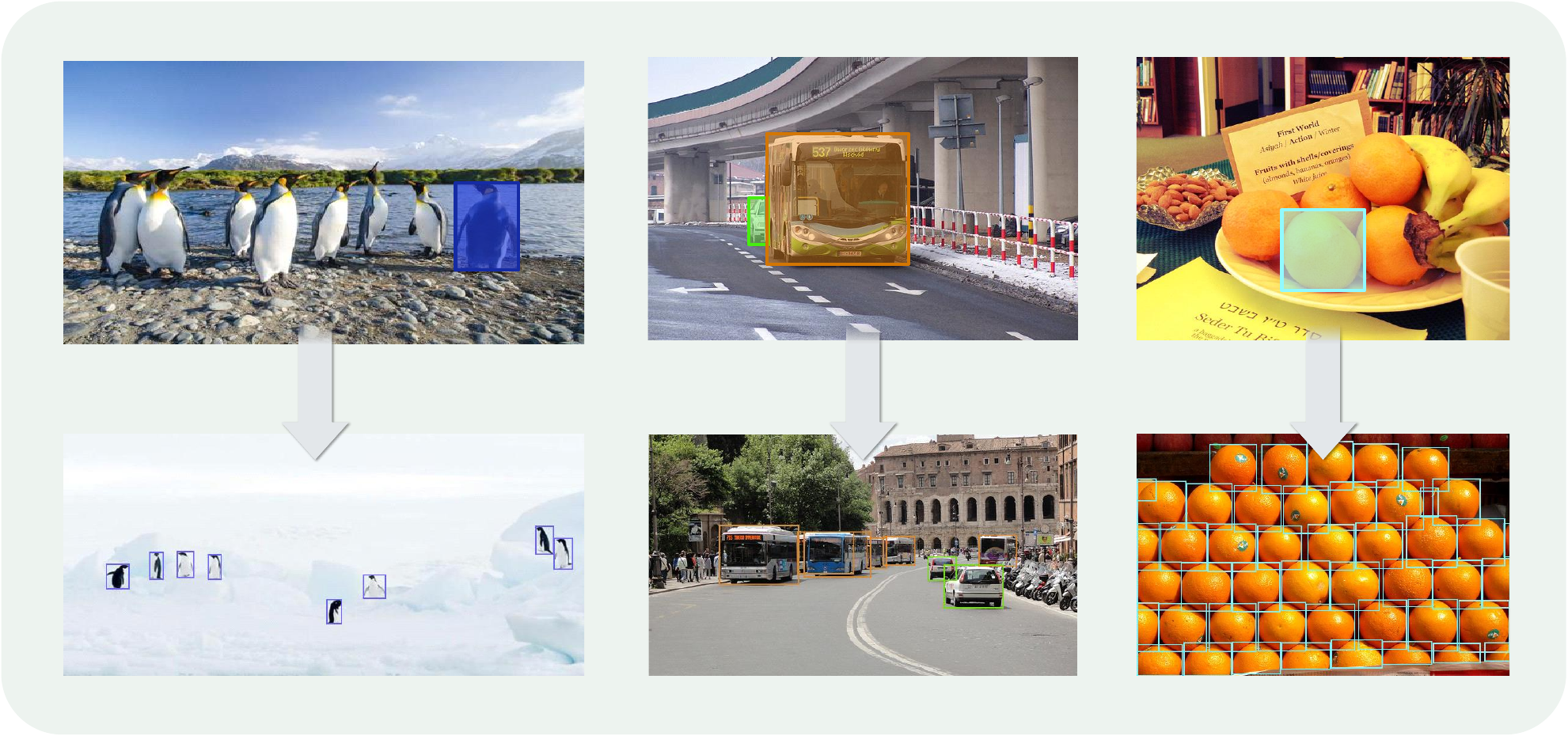} %
\caption{Zero-shot detection visualizations of \textbf{PET-DINO} on \textbf{cross-image exemplar visual} prompt-based detection. Exemplars are shown above, and prediction outputs are shown below.}
\label{visual_I_vis_cross_objects}
\end{figure*}

As illustrated in Figure~\ref{visual_I_vis_cross_objects}, PET-DINO can identify objects of the same category in different images based on a target selected from an exemplar image, paving the way for broader and more general application scenarios.

\begin{figure*}[ht]
\centering
\includegraphics[width=1.0\linewidth]{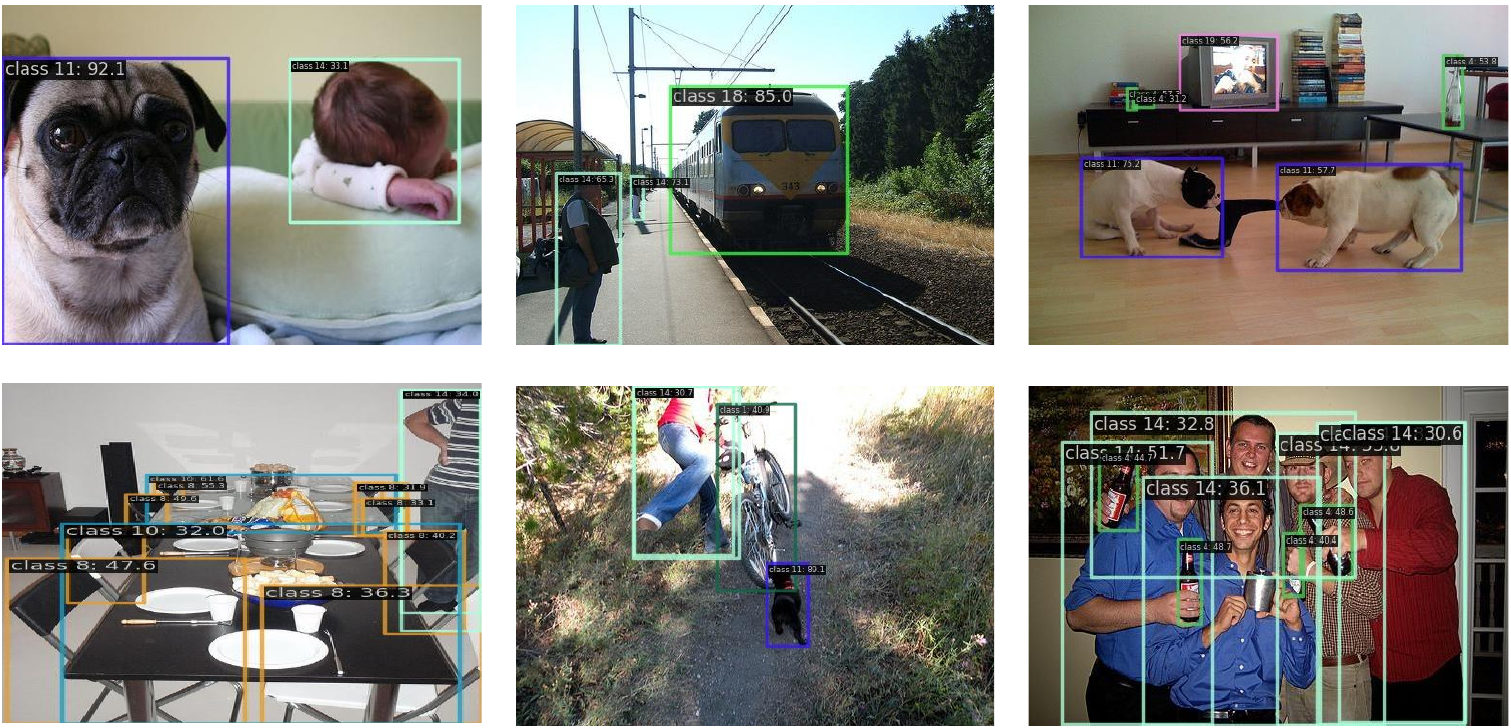} %
\caption{Zero-shot detection visualizations of \textbf{PET-DINO} with \textbf{class-level generic visual} prompts. The visual prompt embeddings are pre-extracted from the training set.}
\label{visual_G_vis}
\end{figure*}

In Figure~\ref{visual_G_vis}, we present zero-shot results based on pre-extracted generic visual prompts. The method achieves strong performance in various scenarios, demonstrating the effectiveness of our training strategies aligned with practical usage. It is especially valuable for adapting to novel scenarios without additional retraining, and can further be applied to cases where textual representations are difficult to align with the targets.

\begin{figure*}[ht]
\centering
\includegraphics[width=1.0\linewidth]{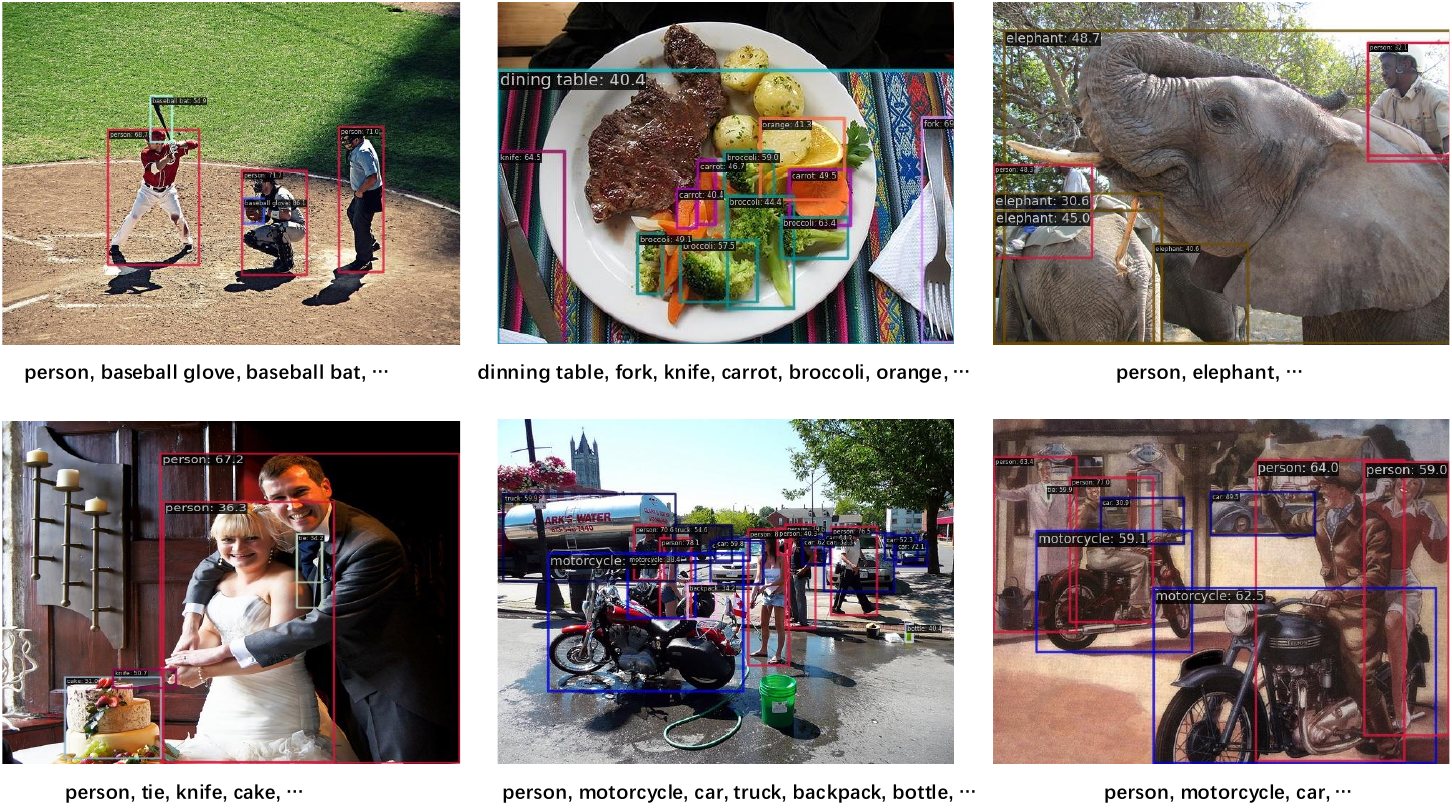} %
\caption{Zero-shot detection visualizations of \textbf{PET-DINO} with \textbf{text} prompts. The category names from the dataset are utilized as textual inputs for prompt generation.}
\label{text_vis}
\end{figure*}

In Figure~\ref{text_vis}, we present zero-shot text prompt-based detection results. Our model maintains robust performance and demonstrates effectiveness across various scenarios, reflecting the harmony of our inheritance strategy.

In conclusion, empowered by our strategies, PET-DINO effectively adapts to diverse application scenarios, exhibiting high accuracy and robust generalization.

\end{document}